\pdfoutput=1

\documentclass[11pt]{article}

\usepackage[preprint]{acl}

\usepackage{times}
\usepackage{latexsym}

\usepackage[T1]{fontenc}

\usepackage[utf8]{inputenc}

\usepackage{microtype}

\usepackage{inconsolata}

\usepackage{graphicx}
\usepackage[most]{tcolorbox}
\usepackage{keystroke}
\usepackage{menukeys}
\usepackage{cleveref}
\usepackage[cjk]{kotex}
\usepackage{booktabs}
\usepackage{multirow}

\usepackage{xcolor}
\usepackage{xspace}
\def\eg{\emph{e.g}.,\xspace}
\def\ie{\emph{i.e}.,\xspace}

\def\tool{\textbf{\textsc{HERITage}}\xspace}

\title{\tool: An End-to-End Web Platform for \\Processing Korean Historical Documents in Hanja}

\author{
  Seyoung Song$^{\diamond}$,
  Haneul Yoo$^{\diamond}$,
  Jiho Jin$^{\diamond}$,
  Kyunghyun Cho$^{\dagger \ddagger}$,
  Alice Oh$^\diamond$
  \\
  \ \\
  $^\diamond$KAIST, $^\dagger$New York University, $^\ddagger$Genentech\\
  \texttt{\{\href{mailto:seyoung.song@kaist.ac.kr}{\color{black}{seyoung.song}}, \href{mailto:haneul.yoo@kaist.ac.kr}{\color{black}{haneul.yoo}}, \href{mailto:jinjh0123@kaist.ac.kr}{\color{black}{jinjh0123}}\}@kaist.ac.kr},\\
  \texttt{kyunghyun.cho@nyu.edu, alice.oh@kaist.edu}
}

\begin{document}
\maketitle
\begin{abstract}
  While Korean historical documents are invaluable cultural heritage, understanding those documents requires in-depth Hanja expertise.
Hanja is an ancient language used in Korea before the 20th century, whose characters were borrowed from old Chinese but had evolved in Korea for centuries.
Modern Koreans and Chinese cannot understand Korean historical documents without substantial additional help, and while previous efforts have produced some Korean and English translations, this requires in-depth expertise, and so most of the documents are not translated into any modern language.
To address this gap, we present \tool, the first open-source Hanja NLP toolkit to assist in understanding and translating the unexplored Korean historical documents written in Hanja.
\tool is a web-based platform providing model predictions of three critical tasks in historical document understanding via Hanja language models: punctuation restoration, named entity recognition, and machine translation (MT).
\tool also provides an interactive glossary, which provides the character-level reading of the Hanja characters in modern Korean, as well as character-level English definition.
\tool serves two purposes.
First, anyone interested in these documents can get a general understanding from the model predictions and the interactive glossary, especially MT outputs in Korean and English.
Second, since the model outputs are not perfect, Hanja experts can revise them to produce better annotations and translations. This would boost the translation efficiency and potentially lead to most of the historical documents being translated into modern languages, lowering the barrier on unexplored Korean historical documents.\footnote{Demo and video are available at \url{https://hanja.dev} and \url{https://hanja.dev/video}}
\end{abstract}

\section{Introduction} %
Korea has one of the largest collections of historical documents in the world, including UNESCO World Heritage texts such as \emph{the Annals of the Joseon Dynasty} (AJD).
The majority of these documents are written in Hanja, which served as a primary writing system in Korea before the 20th century, while modern Korea uses Hangul, a phonetic alphabet, as a writing system from the 20th century until the present day.
Although Hanja is an ancient language whose characters originated from old Chinese characters, it has evolved in Korea, including unique grammatical markers and vocabularies reflecting Korean culture~\cite{handel-2019-sinography, heo-2019-from}.
In other words, modern Koreans or Chinese without Hanja knowledge are unable to read or understand Korean historical documents written in Hanja.
Hence, it is a monumental challenge to comment on and translate these documents into comprehensible modern languages.

\begin{figure}[t!]
  \centering
  \includegraphics[width=\columnwidth]{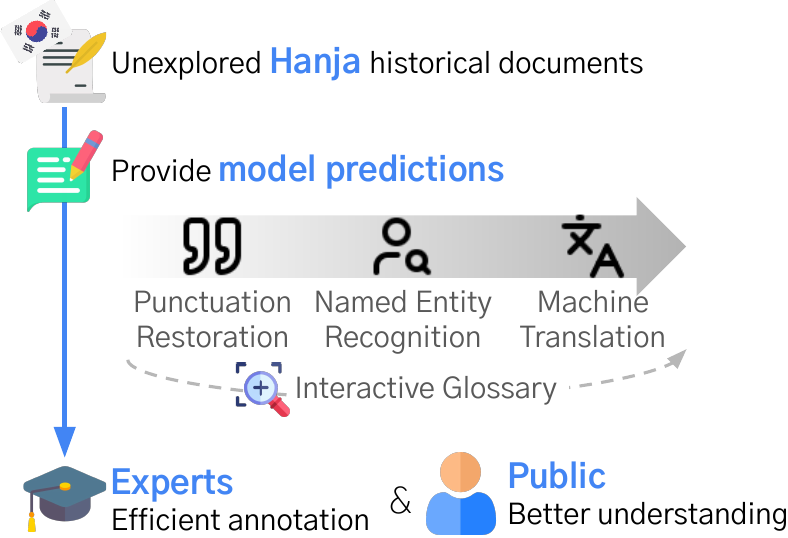}
  \caption{Overview of \tool.}
  \label{fig:teaser_image}
\end{figure}

However, a great bottleneck due to the lack of human resources lies in the Hanja document annotation process.
While approximately 1 billion tokens of Hanja texts are restored and digitized~\cite{classics2021achievements},
only 38\% has been translated into modern Korean in the past 56 years~\citep{song2024does} due to the severe lack of qualified experts, which are fewer than 200 nationwide.
It is noteworthy that training a qualified expert takes up to 7 years, on top of a college degree in Hanja~\cite{kim2021interview}.
For reference, the translation of \emph{the Diaries of the Royal Secretariat} (DRS), which began in 1994, is not projected to be completed by 2048~\cite{park2021seungjeongwon}.

Although natural language processing (NLP) applications have demonstrated their efficacy for historical documents in ancient languages~\cite{sommerschield-etal-2023-machine}, no usable Hanja NLP toolkit has yet to be presented.
Existing solutions trained with insufficient Hanja data achieve poor quality, while state-of-the-art Hanja models on GitHub are inaccessible without computational expertise and resources.
Furthermore, they narrowly focus on specific document types or tasks, not satisfying the needs of the professional annotation workflows (\S\ref{sec:rw}).

To address this gap, we present \tool, a comprehensive web-based platform developed in collaboration with domain experts at the Academy of Korean Studies.
\tool supports three critical tasks in Korean historical document processing in Hanja: punctuation restoration (PR), named entity recognition (NER), and machine translation (MT).
These components mirror and support the traditional expert workflow of punctuation, text collation, vocabulary analysis, and translation~\cite{kim2016ilseongrok}.
Furthermore, interactive glossary and English translations in \tool make Korean historical documents accessible to the global audience for research and education.
We envision \tool 1) to accelerate the annotation and translation process for Hanja experts and 2) to make those documents more accessible to the general public.

Our contributions are as follows:

\begin{itemize}
\item We present \tool, the first Web-based comprehensive platform for AI-assisted processing of Korean historical documents in Hanja.
\item We introduce an end-to-end, integrated Hanja NLP pipeline, comprising punctuation restoration, named entity recognition, and machine translation.
\item We extend the global accessibility of Korean historical documents by providing English translations and a tri-lingual (Hanja, Korean, English) interactive glossary.
\item We publicly release all codebases.\thinspace\footnote{Codes are available at \url{https://github.com/seyoungsong/hanja-platform}}
\end{itemize}

\section{Background: How to Uncover Hanja} %
Hanja, adopted by Korean elites during the Three Kingdoms period (3rd-4th centuries), served as the primary writing system for historical documentation and administration in Korea for over a millennium~\cite{taylor2014writing}.
However, interpreting these documents requires extensive expertise in Hanja, creating a significant barrier to accessibility.
While modern Korean schools continue to teach Hanja (as approximately 66\% of Korean vocabulary consists of Hanja-based words~\cite{heo2010examination}, particularly in academic and technical fields), most students learn only about 900 characters through middle school---a number far insufficient for reading historical texts~\cite{moon2019study}.
This literacy gap has led to a heavy reliance on translated versions for history education, effectively limiting historical perspectives to a single interpretation~\cite{jin2017wicked}.

The conventional workflow of analyzing and translating Hanja documents requires multiple systematic steps~\cite{kim2016ilseongrok}.
First, textual collation is performed to verify the accuracy of the source text by cross-referencing it with related historical documents.
Then, senior Hanja scholars add punctuation marks to segment the text into logical units, as original documents lack modern punctuation.
This is followed by the translation phase, where experts convert the Hanja text into modern languages considering historical context and proper terminology.
Finally, the translation undergoes thorough editorial review, where multiple experts cross-check the accuracy, readability, and adherence to institutional guidelines.
To date, there is no usable NLP toolkit that assists in the exploration and understanding process of Hanja historical documents, which makes the expert annotation process excessively rely on human experts~\cite{park2023ai}.
Publicly available and easily accessible Hanja NLP toolkits will enable 1) an efficient annotation process of Hanja texts for experts and 2) a better general understanding of unexplored Korean historical documents for the public.

\section{\tool} %
We present \tool (Hanja End-to-end platform for Restoration, Identification, Translation), an AI-assisted, web-based platform for processing and analyzing Korean historical documents written in Hanja.
Understanding historical documents requires a systematic approach that mirrors how experts process these texts.
When encountering a new Hanja document, we first need proper text segmentation through punctuation restoration (PR), as original texts lack modern punctuation.
Named entity recognition (NER) then helps identify and verify key proper nouns, which can be cross-referenced with external sources.
Finally, machine translation (MT) provides an initial draft that can be refined using the identified entities and interactive glossary.
While previous research has focused on improving individual tasks, we introduce an integrated pipeline that provides a practical end-to-end solution for processing Hanja documents.

\subsection{System Architecture}

\begin{figure}[tb!]
  \centering
  \includegraphics[width=\linewidth]{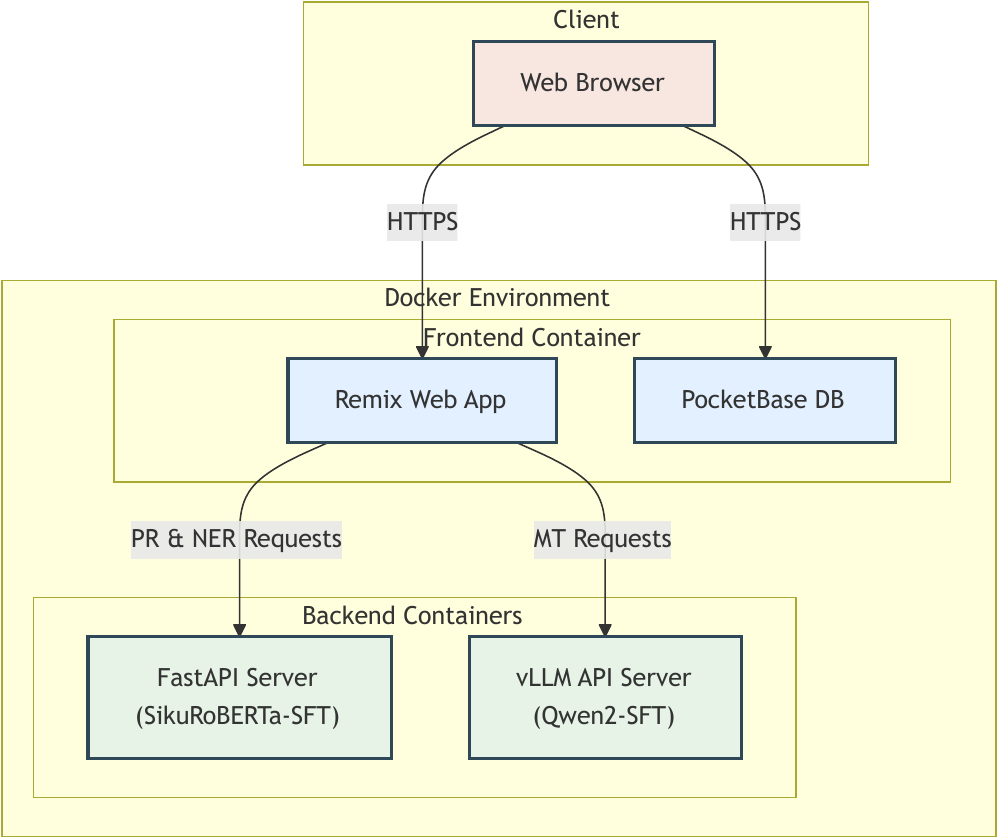}
  \caption{System architecture diagram.}
  \label{fig:system_flow}
\end{figure}

The platform employs a containerized microservice architecture to ensure deployment consistency and scalability (Figure~\ref{fig:system_flow}).
The front-end is a React\thinspace\footnote{\url{https://react.dev}} web application built with the Remix framework\thinspace\footnote{\url{https://remix.run}}, providing task-specific interfaces.
The back-end comprises two primary components: 1) a FastAPI\thinspace\footnote{\url{https://github.com/fastapi/fastapi}} server managing core sequence labeling tasks (punctuation restoration and named entity recognition) using BERT~\cite{devlin-etal-2019-bert}-based models, and 2) a dedicated vLLM\thinspace\footnote{\url{https://github.com/vllm-project/vllm}} instance optimized for efficient machine translation using quantized LLM models.
We use PocketBase\thinspace\footnote{\url{https://pocketbase.io}} for user authentication and session management.

For local deployment, the system requires a modern CPU compatible with Ubuntu 18.04 or higher, Docker installation, and a minimum of 8GB RAM for container operations and BERT model inference.
Machine translation functionality additionally requires a GPU with CUDA 12.1 or higher, compute capability 7.5+, and at least 11GB VRAM.
Our demo website operates on Ubuntu 18.04.6 LTS with an Intel Xeon Silver 4210R CPU (40 cores), 256GB RAM, and 8x NVIDIA GeForce RTX 2080 Ti GPUs (11GB VRAM each).
The platform demonstrates robust performance with processing times of approximately 1 second for PR and NER tasks and up to 10 seconds for MT (approximately 70 tokens per second).
The system can handle at least 50 concurrent users for each task.

\subsection{Web Interface}

The platform comprises a consistent four-panel interface architecture across all tasks: input panel (top), model predictions (left), editable output (right), and interactive glossary (bottom) (Figure~\ref{fig:web_pr}).
This design philosophy treats model outputs as preliminary annotations that require expert validation and refinement.
The key features include:

\begin{figure*}[htb!]
  \centering
  \includegraphics[width=\textwidth]{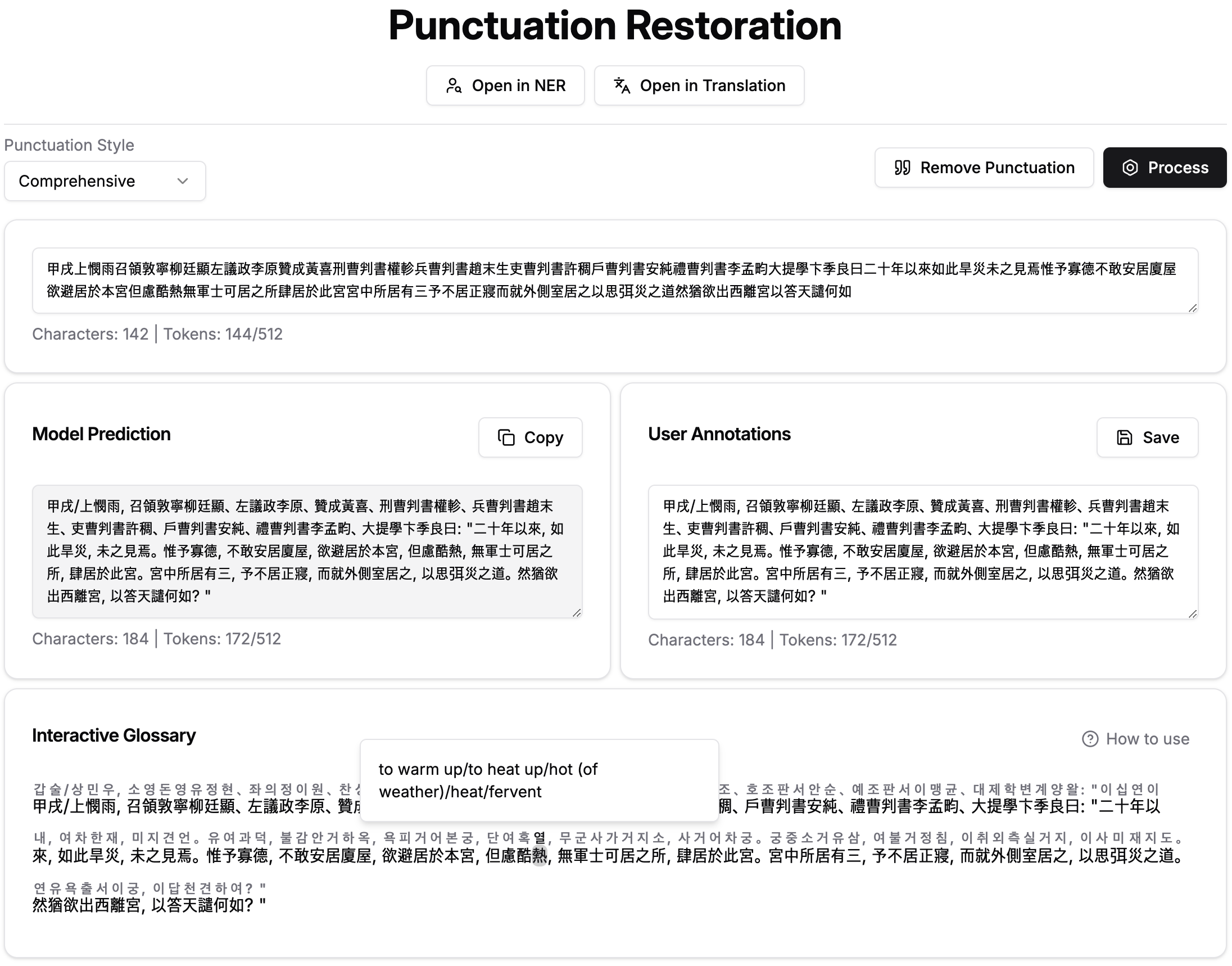}
  \caption{Web interface for Punctuation Restoration.}
  \label{fig:web_pr}
\end{figure*}

\begin{figure*}[htb!]
  \centering
  \includegraphics[width=\textwidth]{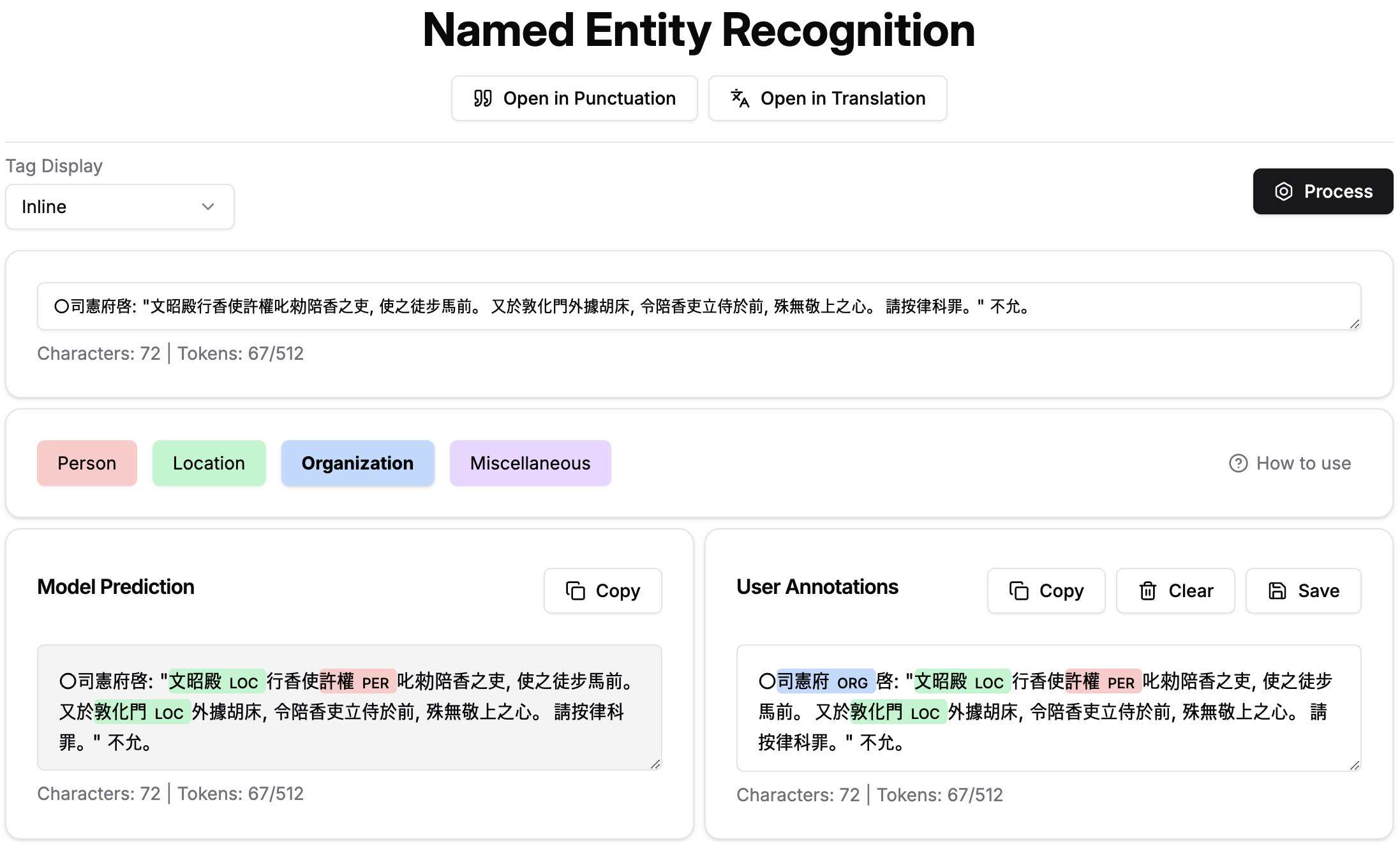}
  \caption{Web interface for Named Entity Recognition.}
  \label{fig:web_ner}
\end{figure*}

\begin{figure*}[htb!]
  \centering
  \includegraphics[width=\textwidth]{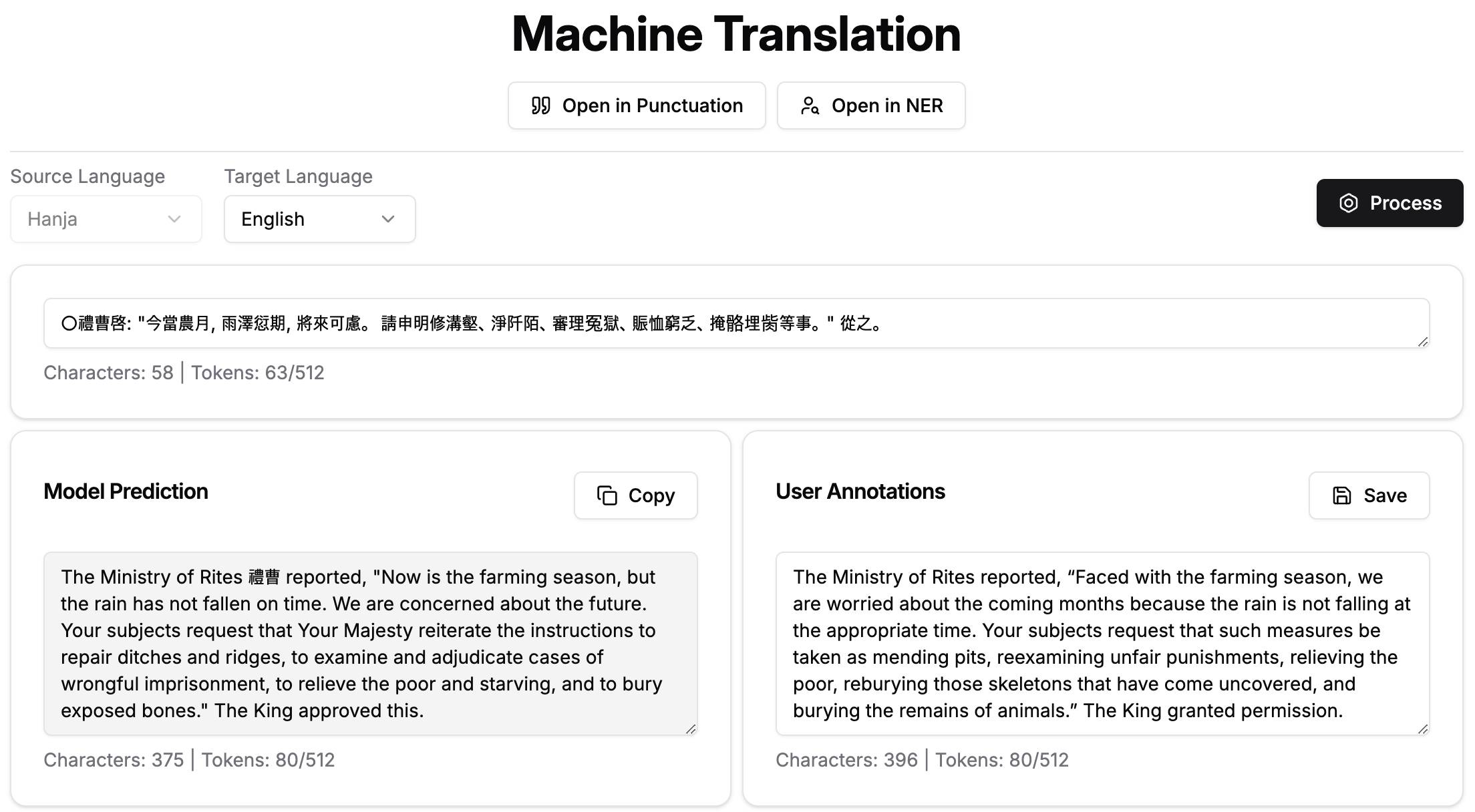}
  \caption{Web interface for Machine Translation.}
  \label{fig:web_mt}
\end{figure*}

\begin{itemize}
\item \textbf{Punctuation Restoration Interface} supports triple annotation modes---1)~\texttt{Comprehensive} (\ie including quotation marks and parentheses), 2) \texttt{Simple} (\ie limited to conventional comma, period, and question marks), and 3) \texttt{Simple w/ space} (\ie whitespace after punctuation marks)---reflecting both modern and traditional punctuation styles (Figure~\ref{fig:web_pr}).
\item \textbf{Named Entity Recognition Interface} features entity visualization through color-coded highlighting with four entity types (\ie \texttt{Person}, \texttt{Location}, \texttt{Organization}, and \texttt{Misc.}) (Figure~\ref{fig:web_ner}). The interface supports both drag-to-tag and click-to-remove operations, with configurable tag display options (\ie hidden, inline, and floating).
\item \textbf{Machine Translation Interface} delivers parallel bilingual output (\ie Hanja-to-English and Hanja-to-Korean) with real-time streaming capabilities to minimize perceived latency during translation (Figure~\ref{fig:web_mt}).
\item \textbf{Interactive Glossary:} displays Korean pronunciations\thinspace\footnote{\url{https://github.com/suminb/hanja}} above Hanja characters using Ruby-style annotations, providing hover-triggered English definitions using CC-CEDICT\thinspace\footnote{\url{https://cc-cedict.org/wiki}} and hyperlinks to an external dictionary\thinspace\footnote{\url{https://hanja.dict.naver.com}} (Figure~\ref{fig:web_pr}).
\end{itemize}

The platform also includes comprehensive user management features, which enable profile customization, annotation history tracking, and data export functionality in both JSON and Excel formats.

\subsection{NLP Models}

We compare seven different models following the methodology of \citet{song2024does}, selecting the best performing model for each task.
Our models are fine-tuned primarily on two corpora: \emph{the Annals of the Joseon Dynasty} (AJD) for historical records by the royal court, and \emph{the Korean Literary Collections} (KLC), which comprises literary works by individual scholars.
We validate the model performances on additional major historical documents, including \emph{the Diaries of the Royal Secretariat} (DRS) and \emph{the Daily Records of the Royal Court and Important Officials} (DRRI).
Detailed training data statistics are provided in Table~\ref{tab:train_data}.

\begin{table}[tb]
  \centering
  \begin{tabular}{@{}llrr@{}}
    \toprule
    \textbf{Task}        & \textbf{Document} & \textbf{\# of Samples} & \textbf{\# of Tokens} \\ \midrule
    \multirow{2}{*}{MT}  & AJD           & 331,150                & 241,653,871           \\
    & KLC           & 53,147                 & 109,406,346           \\ \midrule
    \multirow{2}{*}{NER} & AJD           & 293,854                & 80,841,316            \\
    & KLC           & 8,035                  & 6,673,763             \\ \midrule
    \multirow{2}{*}{PR}  & AJD           & 293,746                & 81,095,372            \\
    & KLC           & 14,428                 & 7,983,038             \\ \bottomrule
  \end{tabular}
  \caption{Training data composition used in our models. Token counts are based on \texttt{cl100k\_base} encoding.}
  \label{tab:train_data}
\end{table}

\paragraph{Punctuation Restoration} We employ SikuRoBERTa~\cite{sikubert-2021} and fine-tune it as a sequence labeling task with a 512-token context window.
Following \citet{pogoda-walkowiak-2021-comprehensive-punctuation}, we implement comprehensive punctuation restoration to capture expert annotation patterns.
The model, built using HuggingFace Transformers~\cite{wolf-etal-2020-transformers}, predicts 23 distinct labels encompassing both individual marks (\eg \texttt{Comma}, \texttt{Period}, and \texttt{Question}) and combinations (\eg \texttt{Colon with quotation}).
The model achieves F1 scores of 88.61 on royal records and 87.76 on literary works.

\paragraph{Named Entity Recognition} For identifying key historical entities~\cite{ehrmann2023named}, we fine-tune SikuRoBERTa to recognize three entity types (\texttt{Person}, \texttt{Location}, and \texttt{Misc.})\footnote{The \texttt{Organization} entity type is omitted due to insufficient training data}.
The model employs IOB2-tagging with a 512-token context window and demonstrates strong performance with F1 scores of 97.53 on royal records and 83.55 on literary works.

\paragraph{Machine Translation} We fine-tune Qwen2-7B~\cite{yang2024qwen2} using LlamaFactory~\cite{zheng-etal-2024-llamafactory} with a 512-token context window.
To optimize inference efficiency, we employ vLLM~\cite{kwon2023efficient} with AWQ quantization~\cite{lin2024awq}.
Our model supports both Hanja-to-Korean and Hanja-to-English translation, achieving BLEU scores of 48.97 (Korean) and 33.15 (English) for royal records, and 33.07 (Korean) for literary works.
For both training and inference, we use the following prompt template:

\begin{tcolorbox}[breakable, enhanced, top=1pt, left=1pt, right=1pt, bottom=1pt]
\small{
Translate the following text from <source language> into <target language>. \keys{\return}

<source language>: <source sentence> \keys{\return}

<target language>:
}
\end{tcolorbox}

\section{Related Work} %
\label{sec:rw}
\subsection{NLP for Hanja}
Prior studies have introduced Korean historical documents to the NLP community and developed Hanja language models specifically trained for individual target tasks, while most models are closed.
\citet{bak-oh-2015-five} first introduced \emph{the Annals of the Joseon Dynasty} (AJD) to the NLP community, and \citet{yoo-etal-2022-hue} introduced a benchmark dataset for Hanja understanding (HUE).
\citet{yang-etal-2023-histred} introduced \emph{Yeonhaengnok} to build a relation extraction dataset (HistRED).
\citet{moon2024exploiting, song2024does} incorporated non-royal literary documents (\ie \emph{diary data (Ilok)}, \emph{letter data (Ganchal)}, and \emph{the Korean Literary Collections} (KLC)) to investigate transferability between Hanja and Classical Chinese.

\citet{yoo-etal-2022-hue} presented pre-trained BERT-based models for Hanja and fine-tuned models for named entity recognition.
Several studies developed translation models from Hanja to old Korean~\cite{park2020ancient, kang-etal-2021-restoring}, contemporary Korean, and English~\cite{son-etal-2022-translating}, while none of them are open source.
\citet{song2024does} publicly released state-of-the-art Hanja NLP models for three individual tasks: punctuation restoration, named entity recognition, and machine translation.

\subsection{NLP Tools for Ancient Languages}
Recent studies on machine learning for NLP have explored historical texts written in ancient languages, including but not limited to ancient Greek, Egyptian hieroglyphs, old Chinese, Latin, and Mayan~\cite{sommerschield-etal-2023-machine}.
However, only a few open models have been developed into feasible NLP tools.
For instance, \citet{sandhan-etal-2023-sanskritshala} presented a web-based annotation toolkit for Sanskrit, integrating word tokenizer, morphological tagger, and dependency parser.

Similarly, GJ.cool\thinspace\footnote{\url{https://gj.cool}} provides a web-based AI platform for optical character recognition, punctuation restoration, and machine translation for Classical Chinese texts.
Despite its utility, it is hardly applicable to other ancient Asian languages in the Sinosphere due to linguistic differences and performance degradation~\cite{song2024does}.
The Institute for the Translation of Korean Classics (ITKC) also offers a web-based machine translation service for Hanja\thinspace\footnote{\url{http://aitr.itkc.or.kr}}.
However, it supports only two curated datasets (\emph{the Diaries of the Royal Secretariat} (DRS) and \emph{the Astronomical Classics}) as inputs, which significantly limits practical usage.

\section{Conclusion} %
We present \tool, the first comprehensive web-based platform for processing unexplored Korean historical documents written in Hanja.
It covers the end-to-end process of Hanja text understanding through three crucial tasks mimicking the workflow of expert annotations: punctuation restoration (PR), named entity recognition (NER), and machine translation (MT).
In addition, it lowers the linguistic barrier of Korean historical documents to international readers by providing English translations and an interactive glossary.
Through its integrated approach, \tool advances digital humanities by democratizing access to Korean historical documents that would otherwise take decades to process.
We believe our open-source platform will accelerate the analysis and translation of Korean historical documents, making centuries of cultural heritage accessible to both scholars and the public.

\section*{Broader Impact Statement}
We have developed a system for efficiently constructing a dataset of an extinct ancient language, Hanja.
It is expected to benefit research and education regarding Hanja and Korean history, and we explicitly prohibit its use for any purposes that could have harmful impacts.
The AI models used in our system have been trained using open-source data and models, and we release all the codebases used in developing the system to contribute to the progress of the broader research community.
We acknowledge that the AI models in this system are imperfect and will explicitly advise users to exercise caution regarding potential errors that may arise during the annotation process.
Furthermore, we are committed to actively incorporating feedback from users and the community to improve the usability and ethical aspects of our system.


\begin{thebibliography}{30}
\providecommand{\natexlab}[1]{#1}

\bibitem[{Bak and Oh(2015)}]{bak-oh-2015-five}
JinYeong Bak and Alice Oh. 2015.
\newblock \href {https://doi.org/10.18653/v1/W15-3702} {Five centuries of
  monarchy in {K}orea: Mining the text of the annals of the {J}oseon dynasty}.
\newblock In \emph{Proceedings of the 9th {SIGHUM} Workshop on Language
  Technology for Cultural Heritage, Social Sciences, and Humanities
  ({L}a{T}e{CH})}, pages 10--14, Beijing, China. Association for Computational
  Linguistics.

\bibitem[{Devlin et~al.(2019)Devlin, Chang, Lee, and
  Toutanova}]{devlin-etal-2019-bert}
Jacob Devlin, Ming-Wei Chang, Kenton Lee, and Kristina Toutanova. 2019.
\newblock \href {https://doi.org/10.18653/v1/N19-1423} {{BERT}: Pre-training of
  deep bidirectional transformers for language understanding}.
\newblock In \emph{Proceedings of the 2019 Conference of the North {A}merican
  Chapter of the Association for Computational Linguistics: Human Language
  Technologies, Volume 1 (Long and Short Papers)}, pages 4171--4186,
  Minneapolis, Minnesota. Association for Computational Linguistics.

\bibitem[{Ehrmann et~al.(2023)Ehrmann, Hamdi, Pontes, Romanello, and
  Doucet}]{ehrmann2023named}
Maud Ehrmann, Ahmed Hamdi, Elvys~Linhares Pontes, Matteo Romanello, and Antoine
  Doucet. 2023.
\newblock \href {https://doi.org/10.1145/3604931} {Named entity recognition and
  classification in historical documents: A survey}.
\newblock \emph{ACM Comput. Surv.}, 56(2).

\bibitem[{Handel(2019)}]{handel-2019-sinography}
Zev Handel. 2019.
\newblock \href {https://doi.org/10.1163/9789004352223} {\emph{Sinography: The
  Borrowing and Adaptation of the Chinese Script}}.
\newblock Brill, Leiden, The Netherlands.

\bibitem[{Heo(2010)}]{heo2010examination}
Chul Heo. 2010.
\newblock \href {https://doi.org/10.17963/ccek.2010..34.221} {Examination how
  many using compound of chinese character words and investigate the frequency
  of use by using analysis of modern korean words 1, 2}.
\newblock \emph{Journal of Chinese Characters Education in Korea},
  (34):221--244.

\bibitem[{Heo(2019)}]{heo-2019-from}
Chul Heo. 2019.
\newblock \href {https://doi.org/10.17320/orient.2019..75.147} {From the point
  of view of academic terms, the term ‘han gukgoyuhanja
  (韓國固有漢字)’ is proposed as a way to solve the problem of
  classification and name of ‘han-character system’}.
\newblock \emph{The Oriental Studies}, 75:147--164.

\bibitem[{{ITKC}(2021)}]{classics2021achievements}
{ITKC}. 2021.
\newblock \href {https://doi.org/10.15752/itkc.59..202111.7} {Achievements of
  the database of korean classics and future tasks}.
\newblock \emph{The Journal of Korean Classics}, 59:7 -- 57.

\bibitem[{Jin(2017)}]{jin2017wicked}
Jae-kyo Jin. 2017.
\newblock \href {https://doi.org/10.15752/itkc.50..201712.105} {Wicked problem
  and retranslation for the translation of korean classics}.
\newblock \emph{The Journal of Korean Classics}, 50:105 -- 142.

\bibitem[{Kang et~al.(2021)Kang, Jin, Yang, Jang, Choo, and
  Kim}]{kang-etal-2021-restoring}
Kyeongpil Kang, Kyohoon Jin, Soyoung Yang, Soojin Jang, Jaegul Choo, and
  Youngbin Kim. 2021.
\newblock \href {https://doi.org/10.18653/v1/2021.naacl-main.317} {Restoring
  and mining the records of the {J}oseon dynasty via neural language modeling
  and machine translation}.
\newblock In \emph{Proceedings of the 2021 Conference of the North American
  Chapter of the Association for Computational Linguistics: Human Language
  Technologies}, pages 4031--4042, Online. Association for Computational
  Linguistics.

\bibitem[{Kim(2021)}]{kim2021interview}
Eun-ju Kim. 2021.
\newblock \href
  {https://web.archive.org/web/20241214023943/https://m.blog.naver.com/foredu0813/222357014590}
  {Artificial intelligence automatically translates classical {Hanja} text
  (인공지능, 한문 고전을 자동으로 번역한다)}.
\newblock \emph{Comma, Period}.
\newblock In Korean.

\bibitem[{Kim(2016)}]{kim2016ilseongrok}
Ok-Kyoung Kim. 2016.
\newblock \href {https://doi.org/10.15752/itkc.47..201606.201} {The ilseongrok
  translation project -current state and future tasks}.
\newblock \emph{The Journal of Korean Classics}, 47:201 -- 239.

\bibitem[{Kwon et~al.(2023)Kwon, Li, Zhuang, Sheng, Zheng, Yu, Gonzalez, Zhang,
  and Stoica}]{kwon2023efficient}
Woosuk Kwon, Zhuohan Li, Siyuan Zhuang, Ying Sheng, Lianmin Zheng, Cody~Hao Yu,
  Joseph Gonzalez, Hao Zhang, and Ion Stoica. 2023.
\newblock \href {https://doi.org/10.1145/3600006.3613165} {Efficient memory
  management for large language model serving with pagedattention}.
\newblock In \emph{Proceedings of the 29th Symposium on Operating Systems
  Principles}, SOSP '23, page 611–626, New York, NY, USA. Association for
  Computing Machinery.

\bibitem[{Lin et~al.(2024)Lin, Tang, Tang, Yang, Chen, Wang, Xiao, Dang, Gan,
  and Han}]{lin2024awq}
Ji~Lin, Jiaming Tang, Haotian Tang, Shang Yang, Wei-Ming Chen, Wei-Chen Wang,
  Guangxuan Xiao, Xingyu Dang, Chuang Gan, and Song Han. 2024.
\newblock \href
  {https://proceedings.mlsys.org/paper_files/paper/2024/file/42a452cbafa9dd64e9ba4aa95cc1ef21-Paper-Conference.pdf}
  {Awq: Activation-aware weight quantization for on-device llm compression and
  acceleration}.
\newblock In \emph{Proceedings of Machine Learning and Systems}, volume~6,
  pages 87--100.

\bibitem[{Moon et~al.(2024)Moon, Kang, Seo, Eo, Park, Yang, and
  Lim}]{moon2024exploiting}
Hyeonseok Moon, Myunghoon Kang, Jaehyung Seo, Sugyeong Eo, Chanjun Park,
  Yeongwook Yang, and Heuiseok Lim. 2024.
\newblock \href {https://doi.org/10.1109/ACCESS.2024.3390181} {Exploiting
  hanja-based resources in processing korean historic documents written by
  common literati}.
\newblock \emph{IEEE Access}, 12:59909--59919.

\bibitem[{Moon et~al.(2019)Moon, Moon, and Park}]{moon2019study}
Younghee Moon, Joon~Hye Moon, and Jiyoung Park. 2019.
\newblock \href {https://doi.org/10.22786/chll.2019..68.007} {A study on the
  selected letters in chinese letter proficiency test}.
\newblock \emph{The Journal of Chinese Language \& Literature}, 68:157 -- 177.

\bibitem[{Park et~al.(2020)Park, Lee, Yang, and Lim}]{park2020ancient}
Chanjun Park, Chanhee Lee, Yeongwook Yang, and Heuiseok Lim. 2020.
\newblock \href {https://doi.org/10.1109/ACCESS.2020.3004879} {Ancient korean
  neural machine translation}.
\newblock \emph{IEEE Access}, 8:116617--116625.

\bibitem[{Park(2023)}]{park2023ai}
Na-Yeon Park. 2023.
\newblock \href
  {http://www.dbpia.co.kr/journal/articleDetail?nodeId=NODE11468512} {Ai
  automatic translation program and utilization prospect for the development of
  the korean studies institute}.
\newblock \emph{Humanities Contents}, (69):75--113.

\bibitem[{Park(2021)}]{park2021seungjeongwon}
Sang-hyun Park. 2021.
\newblock \href {https://www.yna.co.kr/view/AKR20211115145500005} {Ministers
  arguing in front of the king? impossible according to the diaries of the
  royal secretariat (신하들이 왕 앞에서
  옥신각신?…'승정원일기' 보면 불가능)}.
\newblock \emph{Yonhap News Agency}.
\newblock In Korean.

\bibitem[{Pogoda and
  Walkowiak(2021)}]{pogoda-walkowiak-2021-comprehensive-punctuation}
Micha{\l} Pogoda and Tomasz Walkowiak. 2021.
\newblock \href {https://doi.org/10.18653/v1/2021.findings-emnlp.393}
  {Comprehensive punctuation restoration for {E}nglish and {P}olish}.
\newblock In \emph{Findings of the Association for Computational Linguistics:
  EMNLP 2021}, pages 4610--4619, Punta Cana, Dominican Republic. Association
  for Computational Linguistics.

\bibitem[{Sandhan et~al.(2023)Sandhan, Agarwal, Behera, Sandhan, and
  Goyal}]{sandhan-etal-2023-sanskritshala}
Jivnesh Sandhan, Anshul Agarwal, Laxmidhar Behera, Tushar Sandhan, and Pawan
  Goyal. 2023.
\newblock \href {https://doi.org/10.18653/v1/2023.acl-demo.10}
  {{S}anskrit{S}hala: A neural {S}anskrit {NLP} toolkit with web-based
  interface for pedagogical and annotation purposes}.
\newblock In \emph{Proceedings of the 61st Annual Meeting of the Association
  for Computational Linguistics (Volume 3: System Demonstrations)}, pages
  103--112, Toronto, Canada. Association for Computational Linguistics.

\bibitem[{Sommerschield et~al.(2023)Sommerschield, Assael, Pavlopoulos,
  Stefanak, Senior, Dyer, Bodel, Prag, Androutsopoulos, and
  de~Freitas}]{sommerschield-etal-2023-machine}
Thea Sommerschield, Yannis Assael, John Pavlopoulos, Vanessa Stefanak, Andrew
  Senior, Chris Dyer, John Bodel, Jonathan Prag, Ion Androutsopoulos, and Nando
  de~Freitas. 2023.
\newblock \href {https://doi.org/10.1162/coli_a_00481} {Machine learning for
  ancient languages: A survey}.
\newblock \emph{Computational Linguistics}, pages 703--747.

\bibitem[{Son et~al.(2022)Son, Jin, Yoo, Bak, Cho, and
  Oh}]{son-etal-2022-translating}
Juhee Son, Jiho Jin, Haneul Yoo, JinYeong Bak, Kyunghyun Cho, and Alice Oh.
  2022.
\newblock \href {https://doi.org/10.18653/v1/2022.findings-emnlp.91}
  {Translating hanja historical documents to contemporary {K}orean and
  {E}nglish}.
\newblock In \emph{Findings of the Association for Computational Linguistics:
  EMNLP 2022}, pages 1260--1272, Abu Dhabi, United Arab Emirates. Association
  for Computational Linguistics.

\bibitem[{Song et~al.(2024)Song, Yoo, Jin, Cho, and Oh}]{song2024does}
Seyoung Song, Haneul Yoo, Jiho Jin, Kyunghyun Cho, and Alice Oh. 2024.
\newblock \href {https://arxiv.org/abs/2411.04822} {When does classical chinese
  help? quantifying cross-lingual transfer in hanja and kanbun}.
\newblock \emph{arXiv preprint arXiv:2411.04822}.

\bibitem[{Taylor and Taylor(2014)}]{taylor2014writing}
Insup Taylor and M.~Martin Taylor. 2014.
\newblock \href {http://digital.casalini.it/9789027269447} {\emph{Writing and
  literacy in Chinese, Korean and Japanese}}.
\newblock Studies in Written Language and Literacy. John Benjamins Publishing
  Company, Amsterdam.

\bibitem[{Wang et~al.(2022)Wang, Liu, Zhu, Liu, Hu, Shen, and
  Li}]{sikubert-2021}
Dongbo Wang, Chang Liu, Zihe Zhu, Jiangfeng Liu, Haotian Hu, Si~Shen, and Bin
  Li. 2022.
\newblock \href
  {https://kns.cnki.net/kcms/detail/44.1306.G2.20210819.2052.008.html}
  {Construction and application of pre-trained models of siku quanshu in
  orientation to digital humanities}.
\newblock \emph{Library Tribune}, 42(06):31--43.

\bibitem[{Wolf et~al.(2020)Wolf, Debut, Sanh, Chaumond, Delangue, Moi, Cistac,
  Rault, Louf, Funtowicz, Davison, Shleifer, von Platen, Ma, Jernite, Plu, Xu,
  Le~Scao, Gugger, Drame, Lhoest, and Rush}]{wolf-etal-2020-transformers}
Thomas Wolf, Lysandre Debut, Victor Sanh, Julien Chaumond, Clement Delangue,
  Anthony Moi, Pierric Cistac, Tim Rault, Remi Louf, Morgan Funtowicz, Joe
  Davison, Sam Shleifer, Patrick von Platen, Clara Ma, Yacine Jernite, Julien
  Plu, Canwen Xu, Teven Le~Scao, Sylvain Gugger, Mariama Drame, Quentin Lhoest,
  and Alexander Rush. 2020.
\newblock \href {https://doi.org/10.18653/v1/2020.emnlp-demos.6} {Transformers:
  State-of-the-art natural language processing}.
\newblock In \emph{Proceedings of the 2020 Conference on Empirical Methods in
  Natural Language Processing: System Demonstrations}, pages 38--45, Online.
  Association for Computational Linguistics.

\bibitem[{Yang et~al.(2024)Yang, Yang, Hui, Zheng, Yu, Zhou, Li, Li, Liu,
  Huang, Dong, Wei, Lin, Tang, Wang, Yang, Tu, Zhang, Ma, Yang, Xu, Zhou, Bai,
  He, Lin, Dang, Lu, Chen, Yang, Li, Xue, Ni, Zhang, Wang, Peng, Men, Gao, Lin,
  Wang, Bai, Tan, Zhu, Li, Liu, Ge, Deng, Zhou, Ren, Zhang, Wei, Ren, Liu, Fan,
  Yao, Zhang, Wan, Chu, Liu, Cui, Zhang, Guo, and Fan}]{yang2024qwen2}
An~Yang, Baosong Yang, Binyuan Hui, Bo~Zheng, Bowen Yu, Chang Zhou, Chengpeng
  Li, Chengyuan Li, Dayiheng Liu, Fei Huang, Guanting Dong, Haoran Wei, Huan
  Lin, Jialong Tang, Jialin Wang, Jian Yang, Jianhong Tu, Jianwei Zhang,
  Jianxin Ma, Jianxin Yang, Jin Xu, Jingren Zhou, Jinze Bai, Jinzheng He,
  Junyang Lin, Kai Dang, Keming Lu, Keqin Chen, Kexin Yang, Mei Li, Mingfeng
  Xue, Na~Ni, Pei Zhang, Peng Wang, Ru~Peng, Rui Men, Ruize Gao, Runji Lin,
  Shijie Wang, Shuai Bai, Sinan Tan, Tianhang Zhu, Tianhao Li, Tianyu Liu,
  Wenbin Ge, Xiaodong Deng, Xiaohuan Zhou, Xingzhang Ren, Xinyu Zhang, Xipin
  Wei, Xuancheng Ren, Xuejing Liu, Yang Fan, Yang Yao, Yichang Zhang, Yu~Wan,
  Yunfei Chu, Yuqiong Liu, Zeyu Cui, Zhenru Zhang, Zhifang Guo, and Zhihao Fan.
  2024.
\newblock \href {https://arxiv.org/abs/2407.10671} {Qwen2 technical report}.
\newblock \emph{arXiv preprint arXiv:2407.10671}.

\bibitem[{Yang et~al.(2023)Yang, Choi, Cho, and Choo}]{yang-etal-2023-histred}
Soyoung Yang, Minseok Choi, Youngwoo Cho, and Jaegul Choo. 2023.
\newblock \href {https://doi.org/10.18653/v1/2023.acl-long.180} {{H}ist{RED}: A
  historical document-level relation extraction dataset}.
\newblock In \emph{Proceedings of the 61st Annual Meeting of the Association
  for Computational Linguistics (Volume 1: Long Papers)}, pages 3207--3224,
  Toronto, Canada. Association for Computational Linguistics.

\bibitem[{Yoo et~al.(2022)Yoo, Jin, Son, Bak, Cho, and Oh}]{yoo-etal-2022-hue}
Haneul Yoo, Jiho Jin, Juhee Son, JinYeong Bak, Kyunghyun Cho, and Alice Oh.
  2022.
\newblock \href {https://doi.org/10.18653/v1/2022.findings-naacl.140} {{HUE}:
  Pretrained model and dataset for understanding hanja documents of {A}ncient
  {K}orea}.
\newblock In \emph{Findings of the Association for Computational Linguistics:
  NAACL 2022}, pages 1832--1844, Seattle, United States. Association for
  Computational Linguistics.

\bibitem[{Zheng et~al.(2024)Zheng, Zhang, Zhang, Ye, and
  Luo}]{zheng-etal-2024-llamafactory}
Yaowei Zheng, Richong Zhang, Junhao Zhang, Yanhan Ye, and Zheyan Luo. 2024.
\newblock \href {https://doi.org/10.18653/v1/2024.acl-demos.38}
  {{L}lama{F}actory: Unified efficient fine-tuning of 100+ language models}.
\newblock In \emph{Proceedings of the 62nd Annual Meeting of the Association
  for Computational Linguistics (Volume 3: System Demonstrations)}, pages
  400--410, Bangkok, Thailand. Association for Computational Linguistics.

\end{thebibliography}

\end{document}